\documentclass[10pt,twocolumn,letterpaper]{article}

\usepackage{iccv}
\usepackage{times}
\usepackage{epsfig}
\usepackage{graphicx}
\usepackage{amsmath}
\usepackage{amssymb}
\usepackage[dvipsnames]{xcolor}
\usepackage{booktabs}
\usepackage{subfig}
\usepackage{mathtools}
\usepackage{authblk}

\def\youcook{YouCook II\xspace}
\def\bert{BERT\xspace}
\def\vbert{VideoBERT\xspace}

\usepackage[pagebackref=true,breaklinks=true,letterpaper=true,colorlinks,bookmarks=false]{hyperref}

\iccvfinalcopy 

\def\iccvPaperID{1714} 

\ificcvfinal\fi
\begin{document}

\title{VideoBERT: A Joint Model for Video and Language Representation Learning}

\author{Chen Sun}
\author{Austin Myers}
\author{Carl Vondrick}
\author{Kevin Murphy}
\author{Cordelia Schmid}
\affil{Google Research}
\twocolumn[{%
\renewcommand\twocolumn[1][]{#1}%
\maketitle
\begin{center}
    \newcommand{\teaserwidth}{\textwidth}
    \vspace{-0.25in}
    \centerline{
    \includegraphics[width=0.95\teaserwidth,clip]{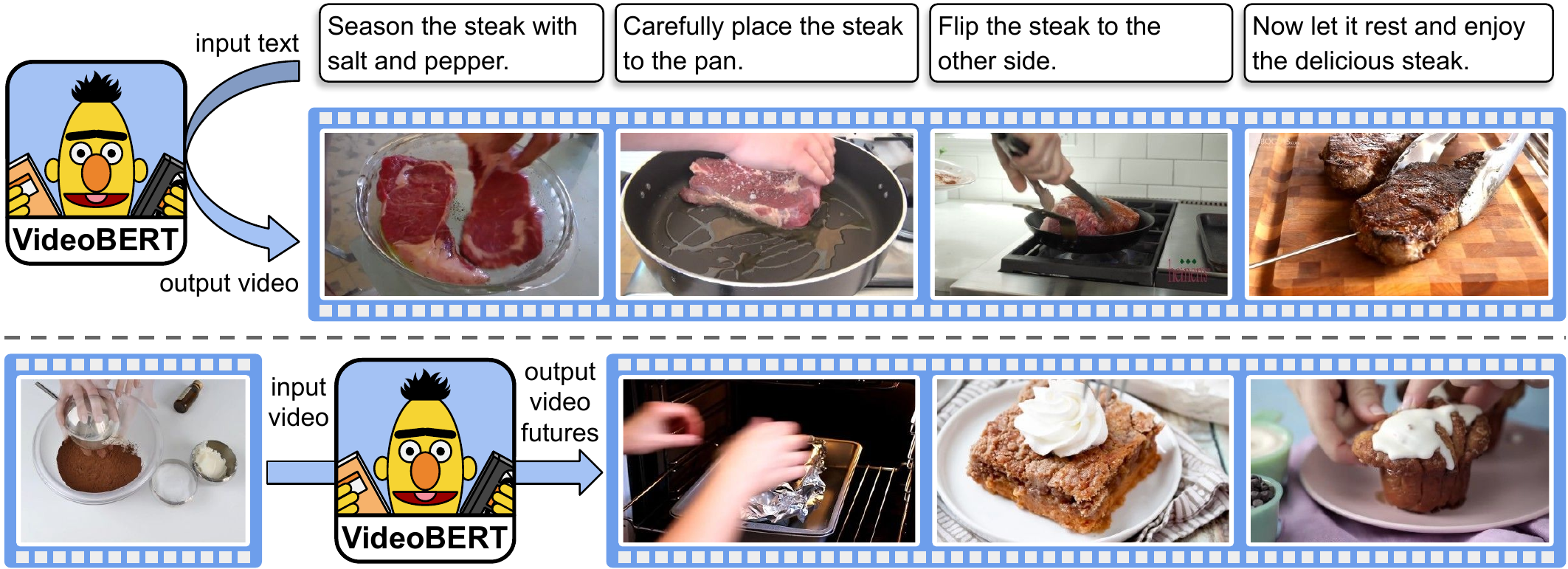}
    }
    \vspace{-0.05in}
    \captionof{figure}{{\bf \vbert text-to-video generation and future forecasting.} 
    (Above) Given some recipe text divided into sentences, $y=y_{1:T}$,
    we generate a sequence of video tokens $x=x_{1:T}$ by computing
    $x_t^* = \arg \max_k p(x_t=k|y)$ using \vbert.
    (Below) Given a video token, 
    we show the top three future tokens forecasted by \vbert at different time scales.
    In this case, \vbert predicts that a bowl of flour and cocoa powder
    may be baked in an oven, and may become a brownie or cupcake.
    We visualize video tokens using the images from the training set closest to centroids in feature space.}
    \vspace{-0.05in}
    \label{fig:teaser}
\end{center}%
}]

\begin{abstract}
\vspace{-0.15in}
Self-supervised learning has become increasingly important 
to leverage the abundance of unlabeled data available on platforms like YouTube.
Whereas most existing approaches learn low-level representations, 
we propose a joint visual-linguistic model to learn high-level features 
without any explicit supervision. 
In particular, inspired by its recent success in language modeling, 
we build upon the BERT model to learn bidirectional joint distributions 
over sequences of visual and linguistic tokens, 
derived from vector quantization of video data 
and off-the-shelf speech recognition outputs, respectively. 
We use \vbert in numerous tasks,
including action classification and video captioning. 
We show that it can be applied directly
to open-vocabulary classification,
and confirm that large amounts of training data 
and cross-modal information are critical to performance.
Furthermore, we outperform the state-of-the-art on video captioning, 
and quantitative results verify that the model learns high-level semantic features.

\end{abstract}
\section{Introduction}

\begin{figure*}[ht]
\centering
\includegraphics[width=.93\textwidth]{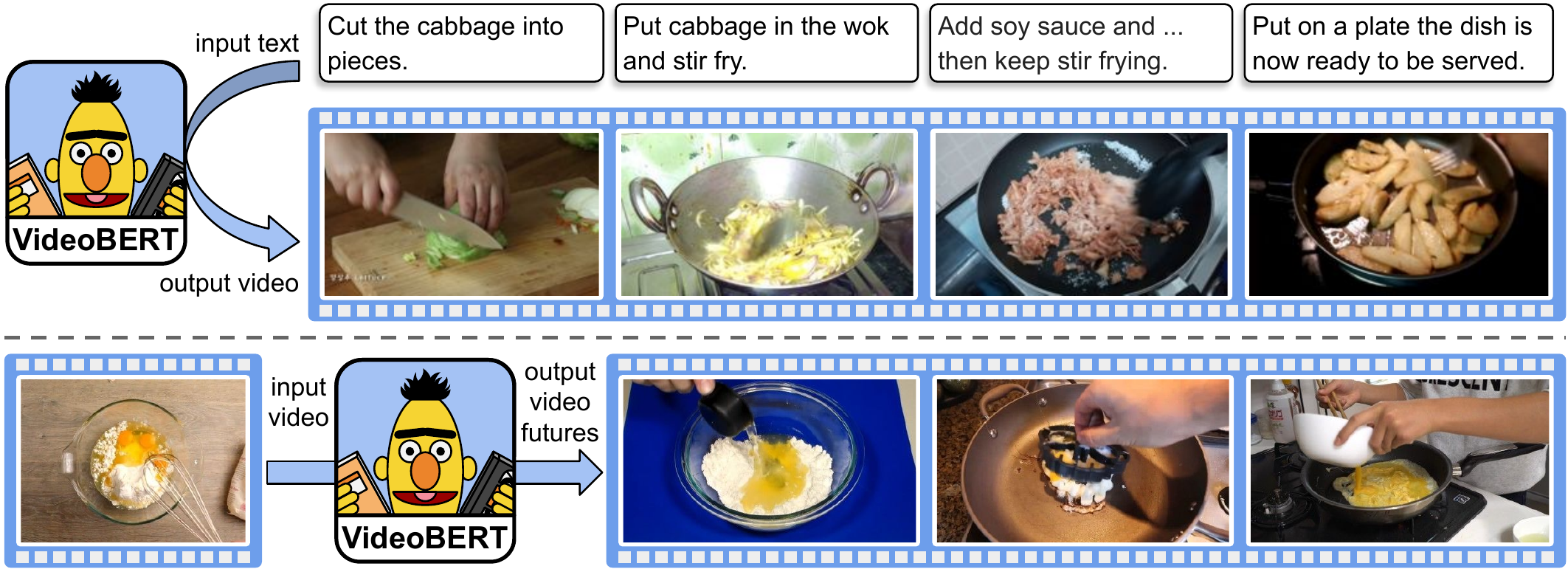}
\caption{
    Additional text-to-video generation and future forecasting examples 
    from \vbert, see Figure~\ref{fig:teaser} for details.
}
\vspace{-0.2in}
\label{fig:teaser_more}
\end{figure*}

Deep learning can benefit a lot from labeled data~\cite{sun2017revisiting}, but this is hard to acquire at scale. Consequently there has been a lot of recent interest in ``self supervised learning'', where we train a model on various ``proxy tasks'', which we hope will result in the discovery of features or representations that can be used in downstream tasks.
A wide variety of such proxy tasks have been proposed in the image and video domains.
However, most of these methods focus on low level features (e.g., textures) and short temporal scales (e.g., motion patterns that last a second or less).
We are interested in discovering high-level semantic features which correspond to actions and events that unfold over longer time scales (e.g.\ minutes), since such representations would be useful for various video understanding tasks.

In this paper, we exploit the key insight that human language 
has evolved words to describe high-level objects and events, 
and thus provides a natural source of ``self'' supervision. 
In particular, we present a simple way to model the relationship 
between the visual domain and the linguistic domain by combining three off-the-shelf methods:
an automatic speech recognition (ASR) system to convert speech into text;
vector quantization (VQ) applied to low-level spatio-temporal visual features derived from pretrained video classfication models;
and the recently proposed \bert model \cite{devlin2018bert}
for learning joint distributions over sequences of discrete tokens.

More precisely, our approach is to apply \bert to learn a model of the form $p(x,y)$, where $x$ is a sequence of ``visual words'', and $y$ is a sequence of spoken words. 
Given such a joint model, we can easily
tackle a variety of interesting tasks.
For example, we can perform text-to-video prediction,
which can be used to automatically illustrate a set of instructions (such as a recipe),
as shown in the top examples of Figure~\ref{fig:teaser} and~\ref{fig:teaser_more}.
We can also perform  the more traditional
video-to-text task of dense video captioning
\cite{Krishna2017} as shown in Figure~\ref{fig:caption_viz}.
In Section~\ref{sec:caption}, we show that our approach
to video captioning significantly
outperforms the previous state-of-the-art~\cite{zhou_captioning_cvpr18} 
on the \youcook dataset~\cite{youcook2}.

We can also use our model in a ``unimodal'' fashion. For example, the implied marginal distribution $p(x)$ is a language model for visual words, which we can use for long-range forecasting. This is illustrated in the bottom examples of Figure~\ref{fig:teaser} and~\ref{fig:teaser_more}.
Of course, there is uncertainty about the future, but the model can generate plausible guesses at a much higher level of abstraction than other deep generative models for video,
such as  those based on VAEs or GANs
(see e.g., 
\cite{Babaeizadeh2018iclr,Denton2018icml,Lee2018arxiv,mocoGAN}),
which tend to predict small changes to low level aspects of the scene, such as the location or pose of a small number of objects.

In summary, our main contribution in this paper is 
a simple way to learn high level video representations 
that capture semantically meaningful and temporally long-range structure. 
The remainder of this paper describes this contribution in detail. 
In particular,
Section~\ref{sec:related} briefly reviews related work;
Section~\ref{sec:method}
describes how we adapt the recent progress in natural language modeling to the video domain;
Section~\ref{sec:results}
presents results on activity recognition and video captioning tasks;
and Section~\ref{sec:concl} concludes.

\section{Related Work}
\label{sec:related}

\textbf{Supervised learning.} 
Some of the most successful approaches for video representation learning 
have leveraged large labeled datasets
(e.g., \cite{kay2017kinetics,monfort2019moments,zhao2017slac,gu2018ava})
to train convolutional neural networks for video classification.
However, it is very expensive to collect such labeled data, and the corresponding label vocabularies are often small and not capable of representing the nuances of many kinds of actions
(e.g., ``sipping'' is slightly different than ``drinking'' which is slightly different than ``gulping'').
In addition,  these approaches are designed for representing short video clips, typically a few seconds long. The main difference to our work is that we focus on the long-term evolution of events in video,
and we do not use manually provided labels.

\textbf{Unsupervised learning.} 
Recently, a variety of approaches for learning density models from video have been proposed.
Some use a single static stochastic variable, 
which is then ``decoded'' into a sequence using an RNN,
either using a VAE-style loss
\cite{Walker2016eccv,Xue2016Visual}
or a GAN-style loss
\cite{vondrick2016generating,mathieu2016deep}.
More recent work uses temporal stochastic variables,
e.g., the SV2P model of 
\cite{Babaeizadeh2018iclr} and the SVGLP model
of \cite{Denton2018icml}.
There are also various GAN-based approaches, such as
the SAVP approach of
\cite{Lee2018arxiv}
and the MoCoGAN approach of \cite{mocoGAN}.
We differ from this work in that we use the BERT model, without any explicit stochastic latent variables,
applied to visual tokens derived from the video.
Thus our model is not a generative model of pixels, but it is a generative model of features derived from pixels,
which is an approach that has been used in other work
(\eg, \cite{Vondrick2016cvpr}).

\textbf{Self-supervised learning.}
To avoid the difficulties of learning a joint model $p(x_{1:T})$,
it has become popular to learn conditional models of the form
$p(x_{t+1:T} | x_{1:t})$, where we partition the signal into two or more blocks, such as gray scale and color, or previous frame and next frame
(e.g., \cite{misra2016shuffle}), and try to predict one from the other (see e.g., \cite{Ranzato2018} for an overview).
Our approach is similar, except we use quantized visual words instead of pixels.
Furthermore, although we learn a set conditional
distributions, our model is a proper joint generative model,
as explained in Section~\ref{sec:method}.

\textbf{Cross-modal learning.} 
The multi-modal nature of video has also been an extensive source of supervision 
for learning video representations, which our paper builds on. 
Since most videos contain synchronized audio and visual signals, 
the two modalities can supervise each other 
to learn strong self-supervised video representations~\cite{aytar2016soundnet,owens2016visually,owens2016ambient}. 
In this work, we use speech (provided by ASR) 
rather than low-level sounds as a source of cross-modal supervision.

\textbf{Natural language models.} 
We build upon recent progress in the NLP community, 
where large-scale language models
such as ELMO \cite{peters2018deep} and \bert \cite{devlin2018bert}
have shown state-of-the-art results for various NLP tasks, 
both at the word level (e.g., POS tagging) and sentence level (e.g., semantic classification). The BERT model is then extended to pre-train on multi-lingual data~\cite{lample2019crosslingual}. Our paper builds on the \bert model to capture structure 
in both the linguistic and visual domains.

\textbf{Image and video captioning.} 
There has been much recent work on image captioning 
(see e.g.,  \cite{kulkarni2011baby,karpathy2015deep,lu2018neural}),
which is a model of the form $p(y|x)$, 
where $y$ is the manually provided caption and $x$ is the image.
There has also been some work on video captioning,
using either manually provided temporal segmentation or estimated
segmentations (see e.g., \cite{Krishna2017,zhou_captioning_cvpr18}).
We use our joint $p(x,y)$ model and apply it to video captioning, 
and achieve state-of-the-art results, 
as we discuss in Section~\ref{sec:caption}.

\textbf{Instructional videos.}
Various papers 
(e.g., \cite{Malmaud2015,alayrac2016unsupervised,Krishna2017,youcook2,zhou_captioning_cvpr18})
have trained models to analyse instructional videos,
such as cooking.
We differ from this work in that we do not use any manual labeling, 
and we learn a large-scale generative model of both words and (discretized) visual signals.

\section{Models}
\label{sec:model}
\label{sec:method}

\begin{figure*}[ht]
\centering
\includegraphics[width=1.0\textwidth]{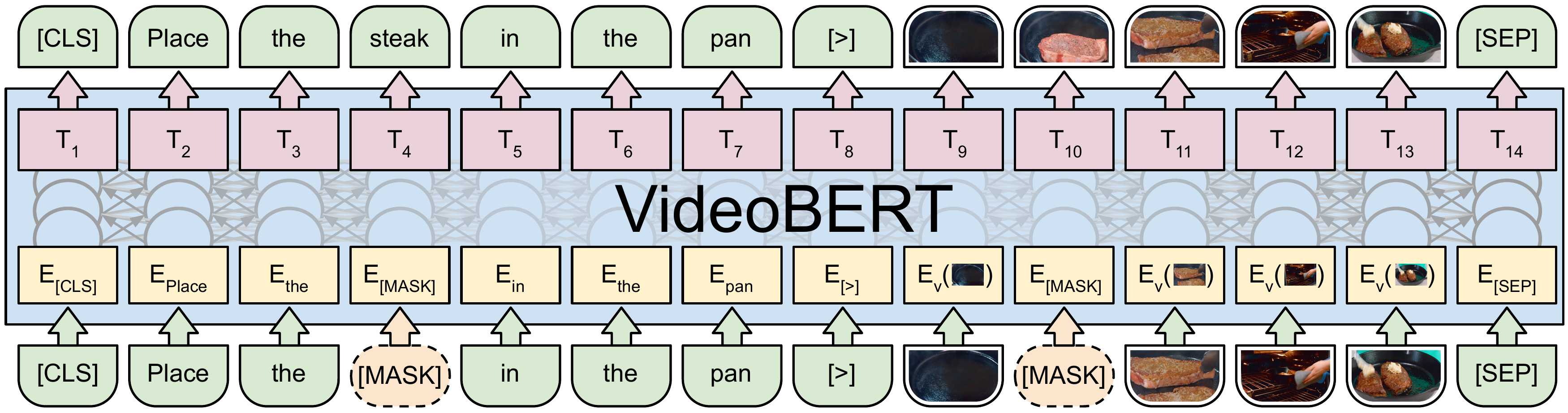}
\caption{
    Illustration of \vbert in the context of a video and text 
    masked token prediction, or {\it cloze}, task. 
    This task also allows for training with text-only and video-only data, 
    and \vbert can furthermore be trained using a 
    linguistic-visual alignment classification objective 
    (not shown here, see text for details).
    }
\label{fig:input_representation}
\end{figure*}

In this section, we briefly summarize the \bert model,
and then describe how we extend it to jointly model video and language data.

\subsection{The \bert model}
\label{sec:bert}

\bert~\cite{devlin2018bert} proposes to learn language representations by using a ``masked language model'' training objective. In more detail, let $x = \{x_1,\ldots,x_L\}$ be a set of discrete tokens, $x_l \in \mathcal{X}$.
We can define a joint probability distribution over this set as follows:
\[
p(x|\theta) = \frac{1}{Z(\theta)} \prod_{l=1}^L \phi_l(x|\theta)
\propto \exp\left( \sum_{l=1}^L \log \phi_l(x|\theta) \right)
\]
where $\phi_{l}(x)$ is the $l$'th potential function,
with parameters $\theta$,
and $Z$ is the partition function.

\newcommand{\xxl}{x_{\setminus l}}

The above model is permutation invariant. In order to capture order information, we can ``tag'' each word with its position in the sentence. 
The \bert model learns an embedding for each of the word tokens, as well as for these tags, and then sums the embedding vectors to get a continuous representation for each token.
The log potential (energy) functions for each location are defined by
\[
\log \phi_l(x|\theta) 
 = x_l^T f_{\theta}(\xxl)
\]
where $x_l$ is a one-hot vector for the $l$'th token (and its tag), and
\[
\xxl = (x_1, \ldots, x_{l-1}, \mathrm{MASK}, x_{l+1}, \ldots, x_L)
\]
The function $f(\xxl)$ is a multi-layer bidirectional transformer model \cite{Vaswani2017} that takes an $L \times D_1$ tensor, containing the $D_1$-dimensional embedding vectors corresponding to $\xxl$,
and returns an $L \times D_2$ tensor,
where $D_2$ is the size of the output of each transformer node.
See \cite{devlin2018bert} for details.
The model is trained to approximately maximize the pseudo log-likelihood
\[
L(\theta) = E_{x \sim D} \sum_{l=1}^L \log p(x_l|\xxl;\theta)
\]
In practice, we can stochastically optimize the logloss (computed from the softmax predicted by the $f$ function) by sampling locations as well as training sentences.

\bert can be extended to model two sentences by concatenating them together.
However, we are often not only interested in simply modeling the extended sequence, 
but rather relationships between the two sentences 
(e.g., is this a pair of consecutive or randomly selected sentences).
\bert accomplishes this by prepending every sequence 
with a special classification token, {\tt \small [CLS]}, 
and by joining sentences with a special separator token, {\tt \small [SEP]}.
The final hidden state corresponding to the {\tt \small [CLS]} token 
is used as the aggregate sequence representation 
from which we predict a label for classification tasks, or which may otherwise be ignored.
In addition to differentiating sentences with the {\tt [SEP]} token, 
\bert also optionally tags each token by the sentence it comes from.
The corresponding joint model can be written as $p(x,y,c)$, 
where $x$ is the first sentence, $y$ is the second, 
and $c = \{0, 1\}$ is a label indicating whether the sentences 
were separate or consecutive in the source document.

For consistency with the original paper,
we also add a {\tt [SEP]} token to the end of the sequence, 
even though it is not strictly needed.
So, a typical masked-out training sentence pair may look like this:
{\tt \small [CLS] let's make a traditional [MASK] cuisine [SEP] orange chicken with [MASK] sauce [SEP]}.
The corresponding class label in this case would be $c=1$,
indicating that $x$ and $y$ are consecutive.

\subsection{The \vbert model}
\label{sec:vbert}

To extend \bert to video,
in such a way that we may still leverage pretrained language models 
and scalable implementations for inference and learning,
we decided to make minimal changes, and transform the raw visual data into a discrete sequence of tokens.
To this end, we propose to generate a sequence of ``visual words'' 
by applying hierarchical vector quantization 
to features derived from the video using a pretrained model.
See Section~\ref{sec:preproc} for details.
Besides its simplicity, this approach encourages the model to focus on 
high level semantics and longer-range temporal dynamics in the video.
This is in contrast to most existing self-supervised approaches to video 
representation learning, which learn low-level properties such as local 
textures and motions, as discussed in Section~\ref{sec:related}.

We can combine the linguistic sentence (derived from the video using ASR)
with the visual sentence to generate data such as this:
{\tt \small 
[CLS] orange chicken with [MASK] sauce [$>$] v01 [MASK] v08 v72 [SEP]},
where {\tt \small v01} and {\tt \small v08} are visual tokens,
and {\tt \small [$>$]} is a special token we introduce 
to combine text and video sentences.
See Figure~\ref{fig:input_representation} for an illustration.

While this {\it cloze} task extends naturally to sequences of linguistic and visual tokens, 
applying a next sentence prediction task, as used by \bert, 
is less straightforward.
We propose a linguistic-visual alignment task, 
where we use the final hidden state 
of the {\tt \small [CLS]} token to predict whether 
the linguistic sentence is temporally aligned with the visual sentence.
Note that this is a noisy indicator of semantic relatedness, 
since even in instructional videos, the speaker may be referring 
to something that is not visually present. 

To combat this, we first randomly concatenate neighboring sentences 
into a single long sentence, 
to allow the model to learn semantic correspondence 
even if the two are not well aligned temporally. 
Second, since the pace of state transitions for even the same action 
can vary greatly between different videos, 
we randomly pick a subsampling rate of 1 to 5 steps for the video tokens.
This not only helps the model be more robust to variations in video speeds,
but also allows the model to capture temporal dynamics
over greater time horizons and learn longer-term state transitions.
We leave investigation into other ways of combining video and text 
to future work.

Overall, we have three training regimes 
corresponding to the different input data modalities: 
text-only, video-only and video-text. 
For text-only and video-only, 
the standard mask-completion objectives are used for training the model.
For text-video, we use the linguistic-visual alignment 
classification objective described above.
The overall training objective 
is a weighted sum of the individual objectives.
The text objective forces \vbert to do well at language modeling;
the video objective forces it to learn a ``language model for video'', 
which can be used for learning dynamics and forecasting;
and the text-video objective forces it to learn 
a correspondence between the two domains.
 
Once we have trained the model, 
we can use it in a variety of downstream tasks,
and in this work we quantitatively evaluate two applications.
In the first application, we treat it as a probabilistic model, 
and ask it to predict or impute the symbols that have been MASKed out.
We illustrate this in Section~\ref{sec:classification},
where we perform ``zero-shot'' classification.
In the second application, we extract the predicted representation 
(derived from the internal activations of the model) for the  {\tt \small [CLS]} token, 
and use that dense vector as a representation of the entire input.
This can be combined with other features derived from the input 
to be used in a downstream supervised learning task.
We demonstrate this in Section~\ref{sec:caption},
where we perform video captioning.

\section{Experiments and Analysis}
\label{sec:results}

In this section we describe our experimental setup, and show quantitative and qualitative results.

\subsection{Dataset}

Deep learning models, in both language and vision domains, 
have consistently demonstrated dramatic gains in performance
with increasingly large datasets. 
For example, the ``large'' \bert model (which we use) 
was pretrained on the concatenation of 
the BooksCorpus (800M words) and English Wikipedia (2,500M words).

Therefore, we would like to train \vbert with a comparably large-scale video dataset.
Since we are interested in the connection between language and vision, 
we would like to find videos where the spoken words are more likely to refer to visual content. 
Intuitively, this is often the case for instructional videos,
and we focus on cooking videos specifically, 
since it is a well studied domain 
with existing annotated datasets available for evaluation.
Unfortunately, such datasets are relatively small, 
so we turn to YouTube to collect a large-scale video dataset for training. 

We extract a set of publicly available cooking videos from YouTube
using the YouTube video annotation system to retrieve videos 
with topics related to ``cooking'' and ``recipe''.
We also filter videos by their duration, 
removing videos longer than 15 minutes, resulting in a set of 312K videos. 
The total duration of this dataset is 23,186 hours, or roughly 966 days.
For reference, this is more than two orders of magnitude larger than 
the next largest cooking video dataset, \youcook, which consists of 
2K videos with a total duration of 176 hours~\cite{youcook2}.

To obtain text from the videos,
we utilize YouTube's automatic speech recognition (ASR) toolkit 
provided by the YouTube Data API~\cite{youtube_api} 
to retrieve timestamped speech information. 
The API returns 
word sequences and the predicted language type. 
Among the 312K videos, ~180K have ASR that can be retrieved by the API, 
and ~120K of these are predicted to be in English. 
In our experiments, 
while we use all videos for the video-only objective,
we only use text from English ASR
for \vbert's text-only and video-text objectives.

We evaluate \vbert on the \youcook dataset~\cite{youcook2},
which contains 2000 YouTube videos averaging 5.26 minutes in duration, for a total of 176 hours. 
The videos have  manually annotated segmentation boundaries and captions. 
On average there are 7.7 segments per video, and 8.8 words per caption.
We use the provided dataset split, with 1333 videos for training and 457 for validation. 
To avoid potential bias during pretraining, 
we also remove any videos which appear in \youcook from our pretraining set.

\subsection{Video and Language Preprocessing}
\label{sec:preproc}

For each input video, we sample frames at 20 fps, 
and create clips from 30-frame (1.5 seconds) non-overlapping windows over the video. 
For each 30-frame clip, we apply a pretrained video ConvNet to extract its features. 
In this work, we use the S3D~\cite{s3dg_2017} which adds separable temporal convolutions 
to an Inception network~\cite{inception} backbone. 
We take the feature activations before the final linear classifier 
and apply 3D average pooling to obtain a 1024-dimension feature vector. 
We pretrain the S3D network on the Kinetics~\cite{kay2017kinetics} dataset, which covers a wide spectrum of actions from YouTube videos, 
and serves as a generic representation for each individual clip.

\begin{figure*}[t]
\centering
\includegraphics[width=1.0\textwidth]{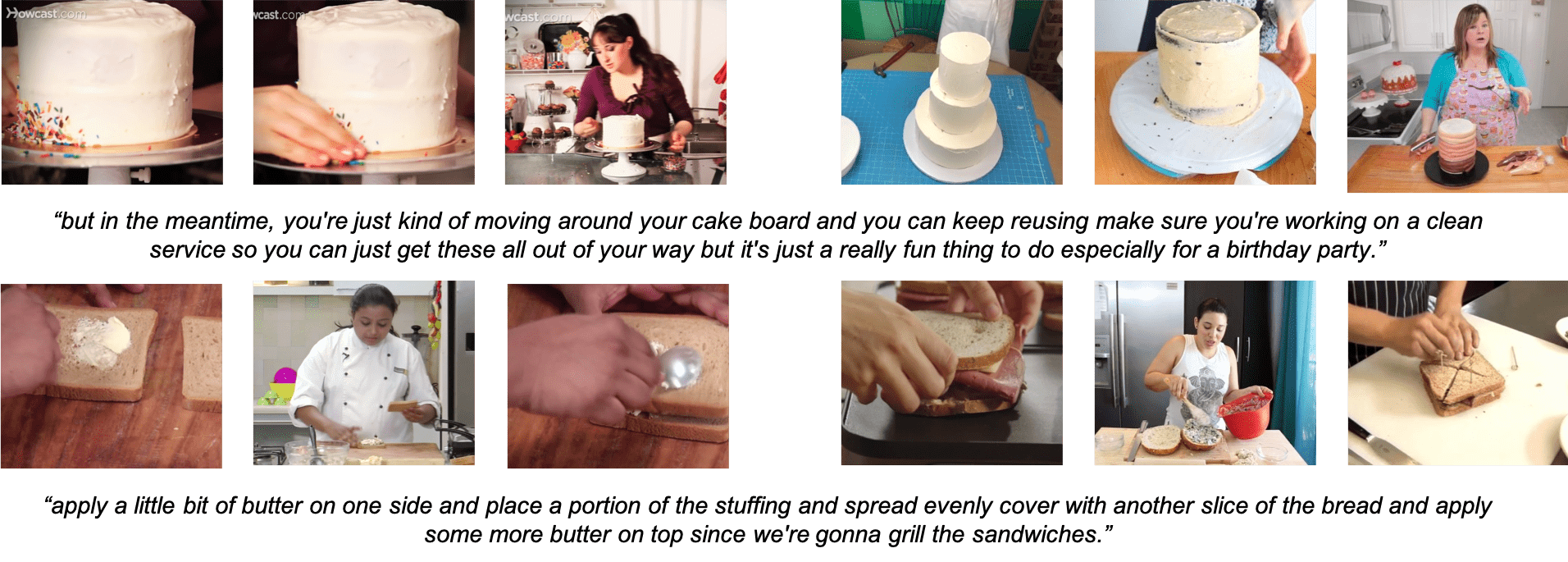}
\caption{Examples of video sentence pairs from the pretraining videos. We quantize each video segment into a token, and then represent it by the corresponding visual centroid. For each row, we show the original frames (left) and visual centroids (right). 
We can see that the tokenization process preserves semantic information
rather than low-level visual appearance.}
\label{fig:pretrain_dataset}
\vspace{-.13in}
\end{figure*}

We tokenize the visual features using hierarchical k-means. 
We adjust the number of hierarchy levels $d$ and the number of clusters per level $k$ 
by visually inspecting the coherence and representativeness of the clusters. 
We set $d\!=\!4$ and $k\!=\!12$, which yields $12^4\! =\! 20736$ clusters in total.
Figure~\ref{fig:pretrain_dataset} illustrates the result of this ``vector quantization'' process.

For each ASR word sequence, we break the stream of words into sentences
by adding punctuation using an off-the-shelf LSTM-based language model.
For each sentence, we follow the standard text preprocessing steps 
from \bert~\cite{devlin2018bert} and tokenize the text into WordPieces~\cite{wordpieces}. 
We use the same vocabulary provided by the authors of \bert, which contains ~30,000 tokens.

Unlike language which can be naturally broken into sentences, 
it is unclear how to break videos into semantically coherent segments. 
We use a simple heuristic to address this problem: 
when an ASR sentence is available,
it is associated with starting and ending timestamps,
and we treat video tokens that fall into that time period as a segment. 
When ASR is not available, we simply treat 16 tokens as a segment.

\subsection{Model Pre-training}

We initialize the \bert weights from a text pre-trained checkpoint. 
Specifically, we use the $\text{BERT}_\text{LARGE}$ model 
released by the authors of~\cite{devlin2018bert},
using the same backbone architecture:
it has 24 layers of Transformer blocks, 
where each block has 1024 hidden units 
and 16 self-attention heads.

We add support for video tokens 
by appending 20,736 entries to the word embedding lookup table 
for each of our new ``visual words''. 
We initialize these entries with the S3D features 
from their corresponding cluster centroids. 
The input embeddings are frozen during pretraining.

Our model training process largely follows the setup of \bert: 
we use 4 Cloud TPUs in the Pod configuration with a total batch size of 128, 
and we train the model for 0.5 million iterations, or roughly 8 epochs. 
We use the Adam optimizer with an initial learning rate of 1e-5, 
and a linear decay learning rate schedule. The training process takes around 2 days.

\subsection{Zero-shot action classification}
\label{sec:classification}

\begin{figure}
\centering
\includegraphics[width=1.0\linewidth]{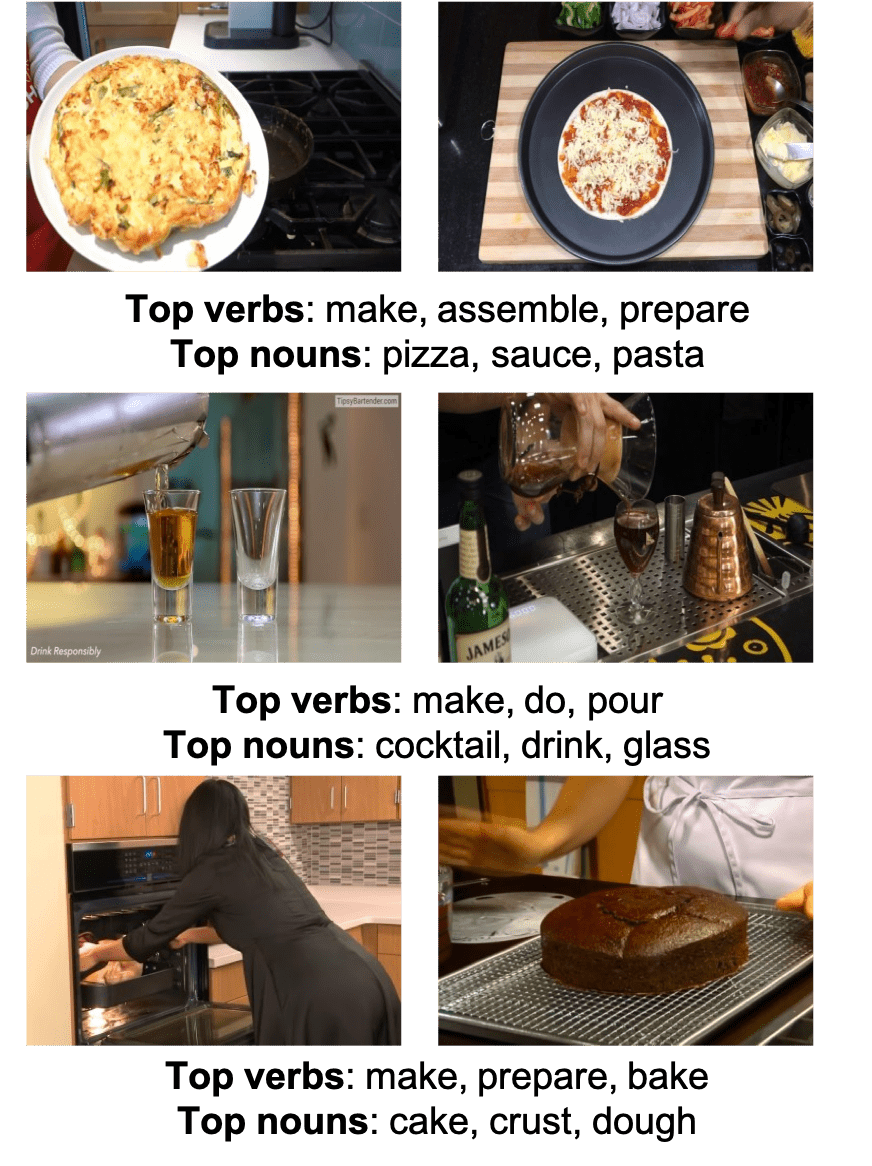}
\caption{Using \vbert to predict nouns and verbs given a video clip. See text for details. The video clip is first converted into video tokens (two are shown here for each example), and then visualized using their centroids.}
\label{fig:qualitative_actions}
\end{figure}

Once pretrained, the \vbert model can be used for 
``zero-shot'' classification on novel datasets, such as \youcook
(By ``zero-shot'' we mean the model is not trained on \youcook data 
nor with the same label ontology used in \youcook).
More precisely, we want to compute $p(y|x)$ 
where $x$ is the sequence visual tokens, 
and $y$ is a sequence of words. 
Since the model is trained to predict sentences, 
we define $y$ to be the fixed sentence,
``{\tt \small now let me show you how to [MASK] the [MASK]},''
and extract the verb and noun labels from the tokens predicted 
in the first and second masked slots, respectively.
See Figure~\ref{fig:qualitative_actions} for some qualitative results.

\begin{table*}

\begin{center}
\scalebox{0.85}{
\begin{tabular}{@{}c|c|c|c|c|c @{}}
\toprule
Method & Supervision & verb top-1 (\%) & verb top-5 (\%) & object top-1 (\%) & object top-5 (\%) \\
\midrule
S3D~\cite{s3dg_2017} & yes & 16.1 & 46.9 & 13.2 & 30.9 \\
\midrule
\bert (language prior) & no & 0.0 & 0.0 & 0.0 & 0.0 \\
\vbert (language prior) & no & 0.4 & 6.9 & 7.7 & 15.3\\
\vbert (cross modal) & no & 3.2 & 43.3 & 13.1 & 33.7\\
\bottomrule
\end{tabular}
}
\end{center}
\vspace{-0.2in}
\caption{Action classification performance on \youcook dataset.
See text for details.
}
\vspace{-0.1in}
\label{tab:youcook_classification_compare}
\end{table*}

\begin{table*}
\begin{center}
\scalebox{0.85}{
\begin{tabular}{@{}c|c|c|c|c|c @{}}
\toprule
Method & Data size & verb top-1 (\%) & verb top-5 (\%) & object top-1 (\%) & object top-5 (\%) \\
\midrule
\vbert & 10K & 0.4 & 15.5 & 2.9 & 17.8\\
\vbert & 50K & 1.1 & 15.7 & 8.7 & 27.3\\
\vbert & 100K & 2.9 & 24.5 & 11.2 & 30.6\\
\vbert & 300K & 3.2 & 43.3 & 13.1 & 33.7\\
\bottomrule
\end{tabular}
}
\end{center}
\vspace{-0.2in}
\caption{Action classification performance on \youcook dataset
as a function of pre-training data size.}
\vspace{-0.2in}
\label{tab:youcook_datasize}
\end{table*}

\begin{table*}
\begin{center}
\scalebox{0.85}{
\begin{tabular}{@{}c|c|c|c|c|c @{}}
\toprule
Method & BLEU-3 & BLEU-4 & METEOR & ROUGE-L & CIDEr  \\
\midrule
Zhou \etal~\cite{zhou_captioning_cvpr18} & 7.53 & 3.84 & 11.55 & 27.44 & 0.38 \\
S3D~\cite{s3dg_2017} & 6.12 & 3.24 & 9.52 & 26.09 & 0.31 \\
\vbert (video only) & 6.33 & 3.81 & 10.81 & 27.14 & 0.47\\
\vbert & 6.80 & 4.04 & 11.01 & 27.50 & 0.49\\
\vbert + S3D & \textbf{7.59} & \textbf{4.33} & \textbf{11.94} & \textbf{28.80} & \textbf{0.55}\\
\bottomrule
\end{tabular}
}
\end{center}
\vspace{-0.1in}
\caption{Video captioning performance on \youcook. We follow the setup from~\cite{zhou_captioning_cvpr18} and report captioning performance on the validation set, given ground truth video segments.
Higher numbers are better.
}
\label{tab:youcook_captioning}
\end{table*}

\begin{figure*}
\centering
\includegraphics[width=.95\linewidth]{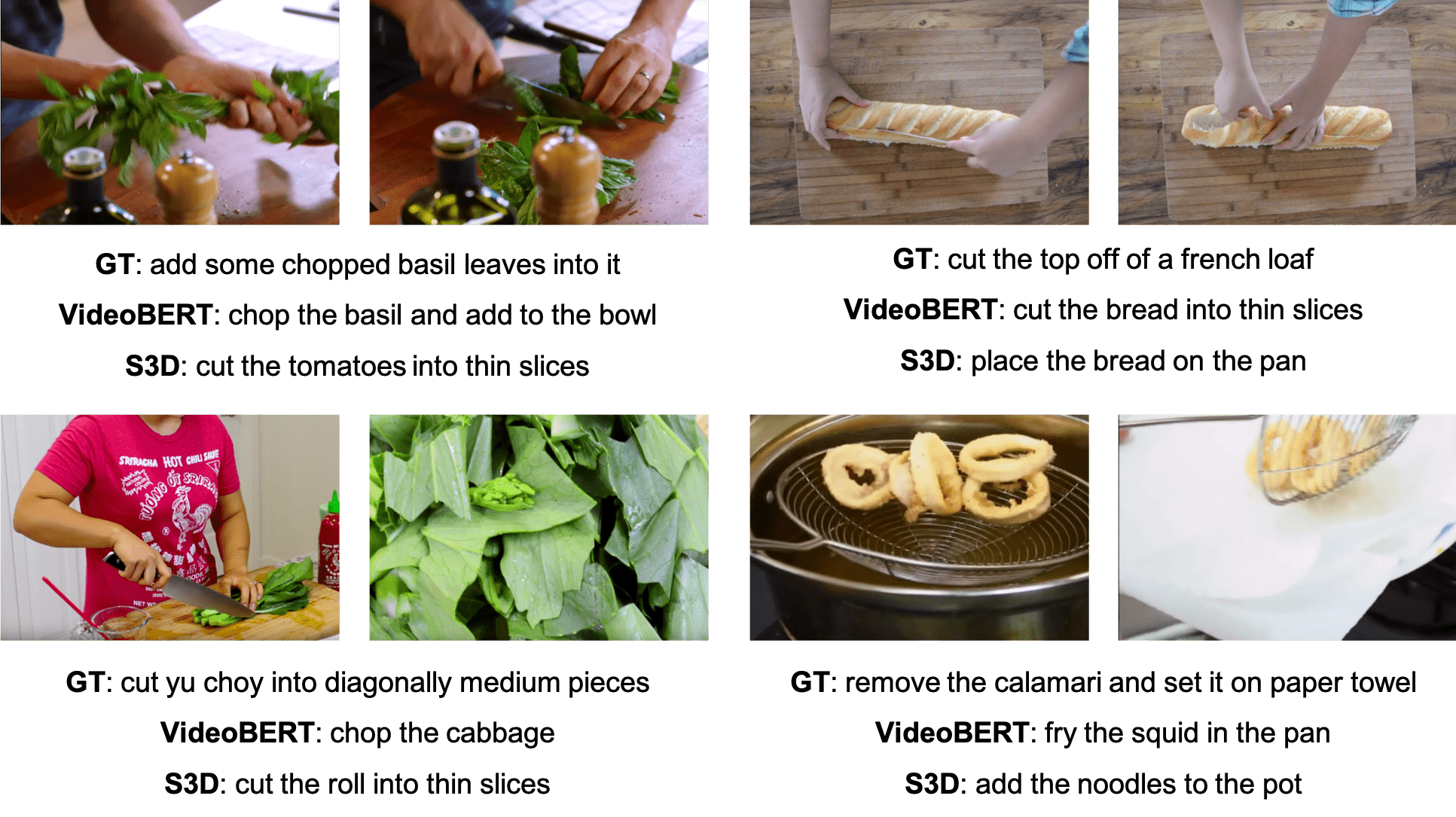}
\caption{Examples of generated captions by \vbert and the S3D baseline.
In the last example, \vbert fails to exploit the full temporal context, since it misses the paper towel frame.}
\vspace{-.1in}
\label{fig:caption_viz}
\end{figure*}

For quantitative evaluation, we use the \youcook dataset.
In~\cite{zhou2018weakly}, the authors collected ground truth bounding boxes 
for the 63 most common objects for the validation set of \youcook. 
However, there are no ground truth labels for actions, 
and many other common objects are not labeled. 
So, we collect action and object labels,
derived from the  ground truth captions, to address this shortcoming. 
We run an off-the-shelf part-of-speech tagger on the ground truth captions 
to retrieve the 100 most common nouns 
and 45 most common verbs, 
and use these to derive ground truth labels.
While \vbert's word piece vocabulary gives it the power to 
effectively perform open-vocabulary classification, 
it is thus more likely to make semantically correct predictions 
that do not exactly match the more limited ground truth. 
So, we report both top-1 and top-5 classification accuracy metrics,
where the latter is intended to mitigate this issue, 
and we leave more sophisticated evaluation techniques for future work.
Lastly, if there is more than one verb or noun associated with a video clip, 
we deem a prediction correct if it matches any of those. 
We report the performance on the validation set of \youcook.

Table~\ref{tab:youcook_classification_compare} 
shows the top-1 and top-5 accuracies of \vbert and its ablations. 
To verify that \vbert actually makes use of video inputs, 
we first remove the video inputs to \vbert, 
and use just the language model $p(y)$ to perform prediction.
We also use the language prior from the text-only \bert model, 
that was not fine-tuned on cooking videos. 
We can see that \vbert significantly outperforms both baselines. 
As expected, the language prior of \vbert is adapted to cooking sentences, 
and is better than the vanilla \bert model.

We then compare with a fully supervised classifier 
that was trained using the training split of \youcook. 
We use  the pre-computed S3D features (same as the inputs to \vbert), 
applying average pooling over time, followed by a linear classifier.
Table~\ref{tab:youcook_classification_compare} shows the results. 
As we can see, the supervised framework outperforms \vbert in top-1 verb accuracy,
which is not surprising given that \vbert has an effectively open vocabulary.
(See Figure~\ref{fig:qualitative_actions} for an illustration 
of the ambiguity of the action labels.)
However, the top-5 accuracy metric reveals that \vbert achieves comparable performance
to the fully supervised S3D baseline, without using any supervision from \youcook, 
indicating that the model is able to perform competitively in this ``zero-shot'' setting.

\subsection{Benefits of large training sets}

We also studied the impact of the size of the pretraining dataset. 
For this experiment, we take random subsets of 10K, 50K and 100K videos 
from the pretraining set, and pretrain \vbert using the same setup as above, 
for the same number of epochs. 
Table~\ref{tab:youcook_datasize} shows the performance.
We can see that the accuracy grows monotonically as the amount of data increases, 
showing no signs of saturation. 
This indicates that \vbert may benefit from even larger pretraining datasets.

\subsection{Transfer learning for captioning}
\label{sec:caption}

We further demonstrate the effectiveness of \vbert when used as a feature extractor. 
To extract features given only video inputs, we again use a simple fill-in-the-blank task, by appending the video tokens to a template sentence ``{\tt \small now let's [MASK] the [MASK] to the [MASK], and then [MASK] the [MASK]}.'' We extract the features for the video tokens and the masked out text tokens, take their average and concatenate the two together, to be used by a supervised model in a downstream task.

We evaluate the extracted features on video captioning, following the setup from~\cite{zhou_captioning_cvpr18}, where the ground truth video segmentations are used to train a supervised model mapping video segments to captions.
We use the same model that they do,
namely a transformer encoder-decoder, 
but we replace the inputs to the 
encoder with the features derived from \vbert described above. 
We also concatenate the \vbert features with average-pooled S3D features; as a baseline, we also consider using just S3D features without \vbert.
We set the number of Transformer block layers to 2, the hidden unit size to 128, and Dropout probability to 0.4.
We use a 5-fold cross validation on the training split to set the hyper-parameters, and report performance on the validation set. 
We train the model for 40K iterations with batch size of 128. 
We use the same Adam optimizer as in \vbert pre-training, and set the initial learning rate to 1e-3 with a linear decay schedule.

Table~\ref{tab:youcook_captioning} shows the results. We follow the standard practice in machine translation and compute BLEU and METEOR scores micro-averaged at corpus level, and also report ROUGE-L~\cite{lin2004rouge} and CIDEr~\cite{vedantam2015cider} scores. For the baseline method~\cite{zhou_captioning_cvpr18}, we recompute the metrics using the predictions provided by the authors. We can see that \vbert consistently outperforms the S3D baseline, especially for CIDEr. We can also see that cross-modal pretraining outperforms the video-only version. Furthermore, by concatenating the features from \vbert and S3D, the model achieves the best performance across all metrics\footnote{The metrics used by~\cite{zhou_captioning_cvpr18} are macro-averaged at video level and may suffer from undesirable sparsity artifacts. Using their provided evaluation code, \vbert + S3D has B@4 of 1.79, and METEOR of 10.80.}.

Figure~\ref{fig:caption_viz} shows some qualitative results. We note that the predicted word sequence is rarely exactly equal to the ground truth, which explains why the  metrics in Table~\ref{tab:youcook_captioning} (which measure n-gram overlap) are all low in absolute value.
However, semantically the results seem reasonable.

\section{Discussion and conclusion}
\label{sec:concl}

This paper adapts the powerful \bert model to learn 
a joint visual-linguistic representation for video.
Our experimental results demonstrate that 
we are able to learn high-level semantic representations, 
and we outperform the state-of-the-art 
for video captioning on the \youcook dataset.
We also show that this model can be used directly for open-vocabulary classification, 
and that its performance grows monotonically with the size of training set.

This work is a first step in the 
direction of learning such joint representations.
For many applications, including cooking,
it is important to use spatially fine-grained visual representations, 
instead of just working at the frame or clip level, 
so that we can distinguish individual objects and their attributes.
We envision either using pretrained object detection and semantic segmentation models, 
or using unsupervised techniques for broader coverage.
We also want to explicitly model visual patterns at multiple temporal scales, 
instead of our current approach, that skips frames but builds a single vocabulary.

Beyond improving the model,
we plan to assess our approach on 
other video understanding tasks,
and on other domains besides cooking.
(For example, we may use the recently released COIN dataset of manually labeled instructional videos \cite{Tang2019}.)
We believe the future prospects for large scale representation learning from video and language look quite promising.
\vspace{-0.1in}

\vspace{0.2in}
\noindent {\bf Acknowledgements.}
We would like to thank Jack Hessel, Bo Pang, Radu Soricut, Baris Sumengen, Zhenhai Zhu, and the BERT team for sharing amazing tools that greatly facilitated our experiments; Justin Gilmer, Abhishek Kumar, David Ross, and Rahul Sukthankar for helpful discussions. Chen would like to thank Y. M. for inspiration.


{\small
\bibliographystyle{ieee_fullname}
\bibliography{egbib}
}

\newpage
\renewcommand{\thefigure}{A\arabic{figure}}
\setcounter{figure}{0}
\begin{figure*}
    \centering
    \includegraphics[width=0.99\linewidth]{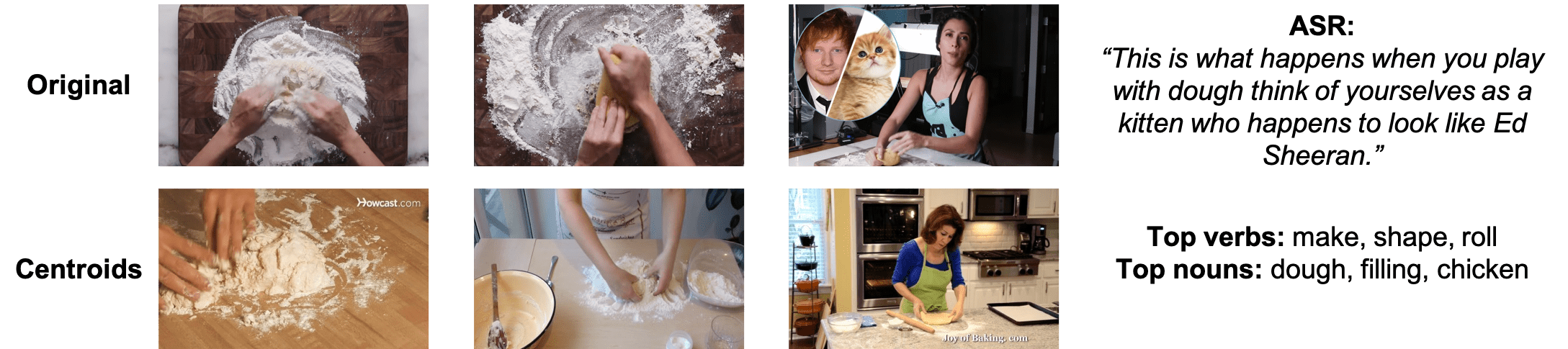}\vspace{20pt}
    \includegraphics[width=0.99\linewidth]{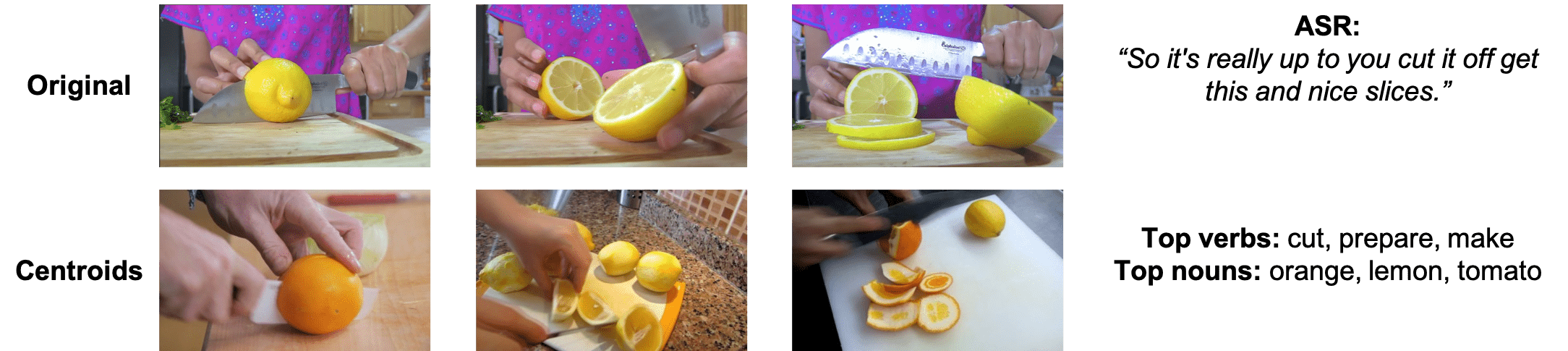}\vspace{20pt}
    \includegraphics[width=0.99\linewidth]{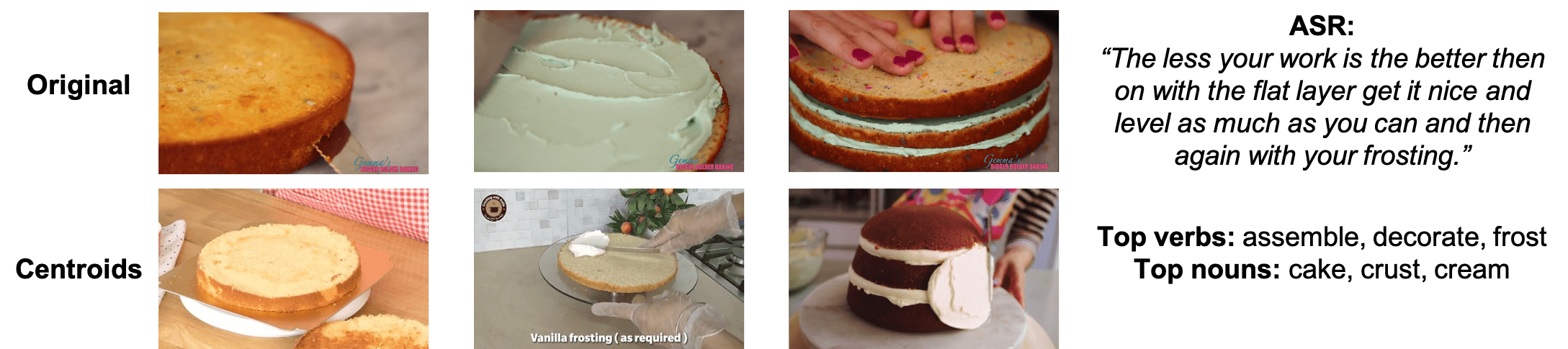}\vspace{20pt}
    \includegraphics[width=0.99\linewidth]{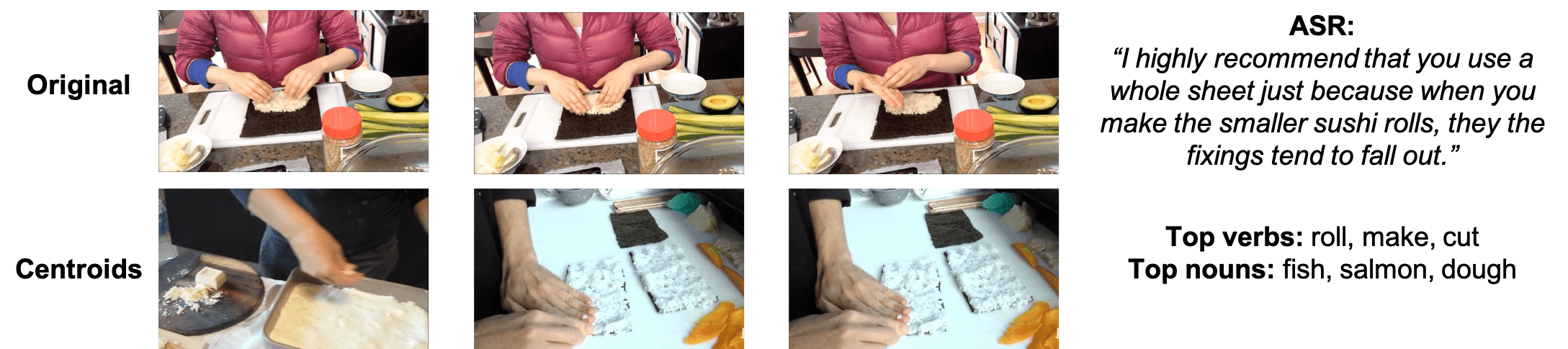}
    \caption{Visualizations for video to text prediction. For each example, we show the key frames from the original video (top left) and the associated ASR outputs (top right), we then show the centroid images of video tokens (bottom left) and the top predicted verbs and nouns by \vbert (bottom right). Note that the ASR outputs are not used to predict verbs and nouns.}
\end{figure*}

\begin{figure*}
    \centering
    \includegraphics[width=0.95\linewidth]{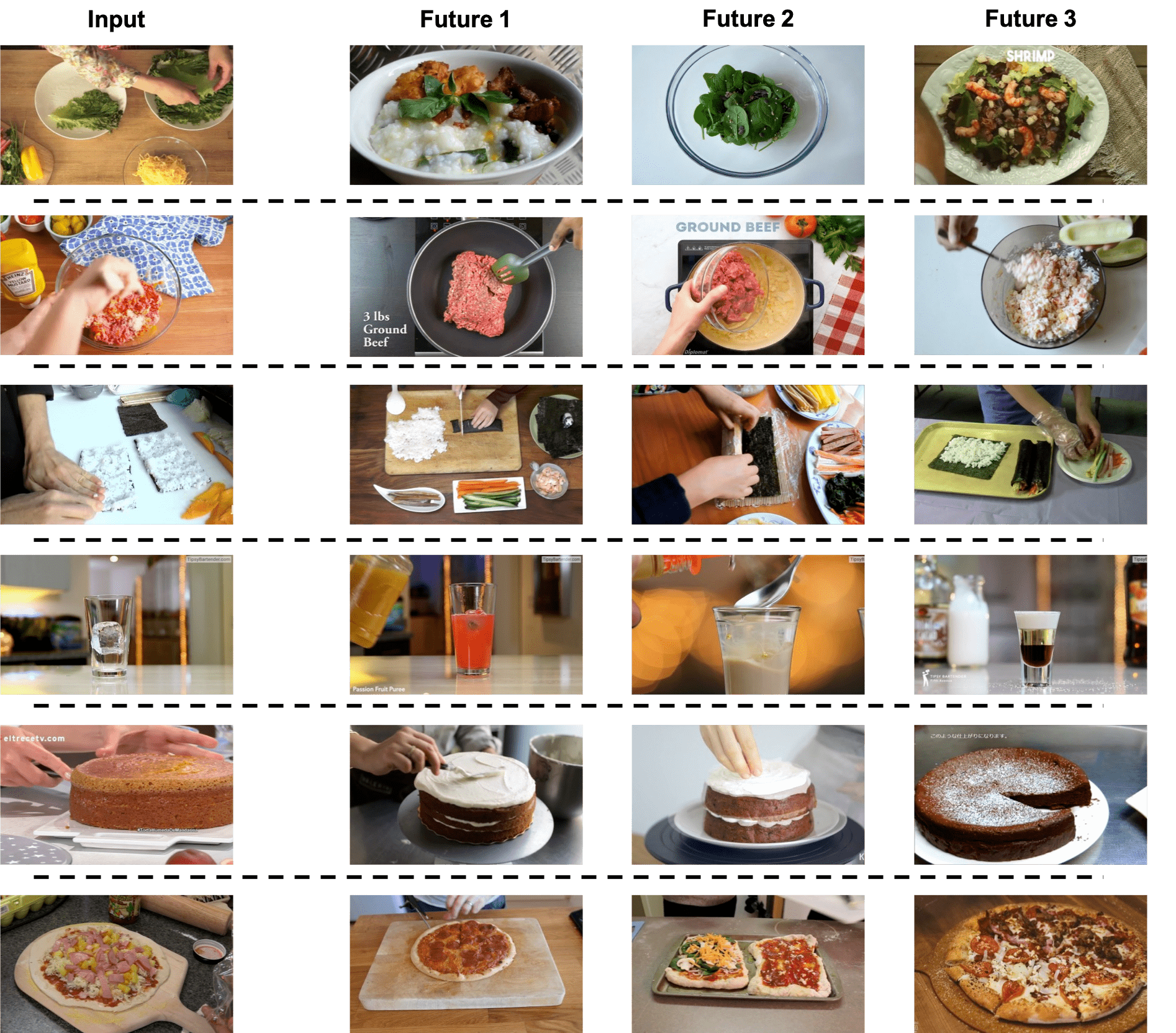}
    \caption{Visualizations for video to video prediction. Given an input video token, we show the top 3 predicted video tokens 2 steps away in the future. We visualize each video token by the centroids.}
\end{figure*}

\begin{figure*}
    \centering
    \includegraphics[width=0.75\linewidth]{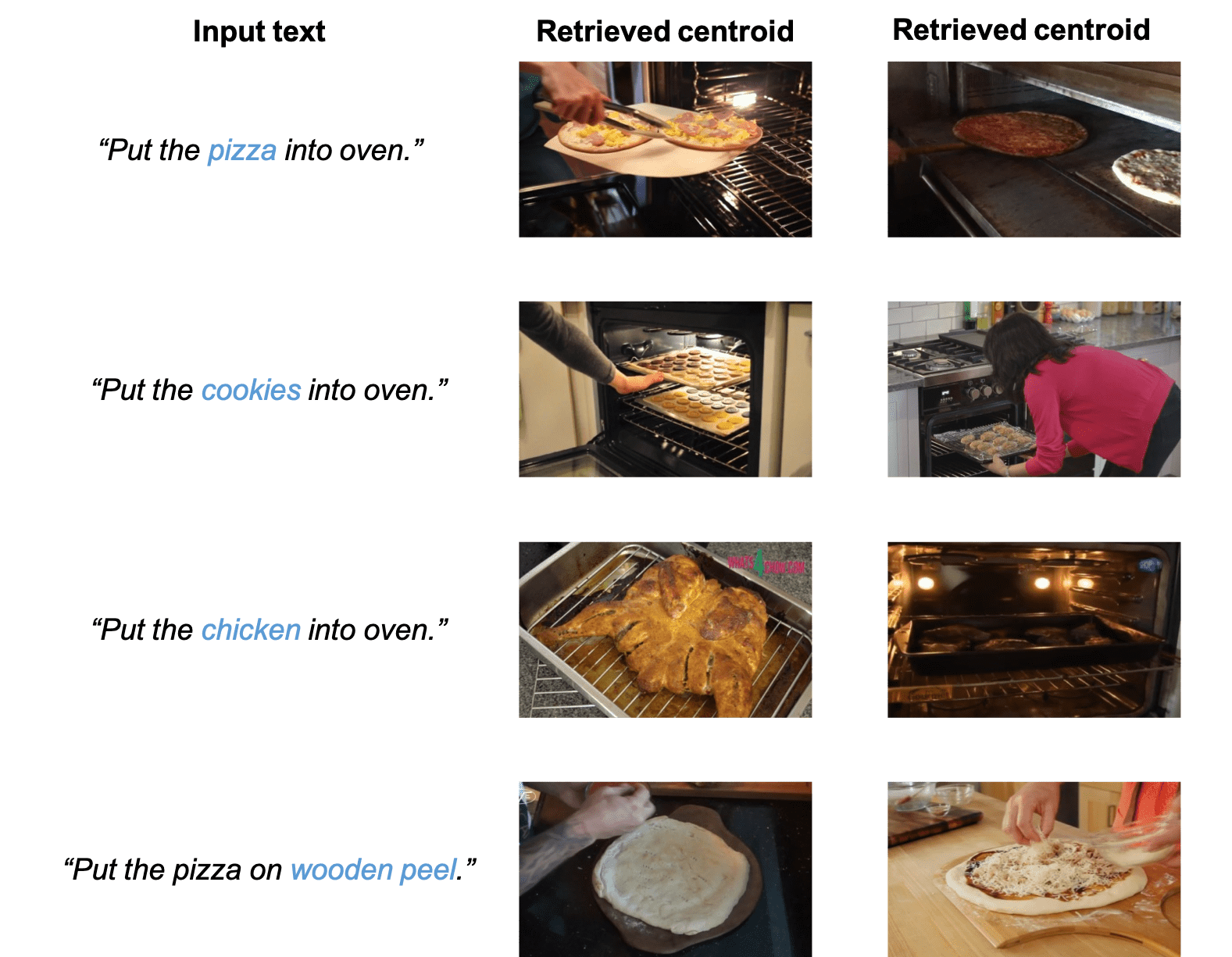}\vspace{25pt}
    \includegraphics[width=0.75\linewidth]{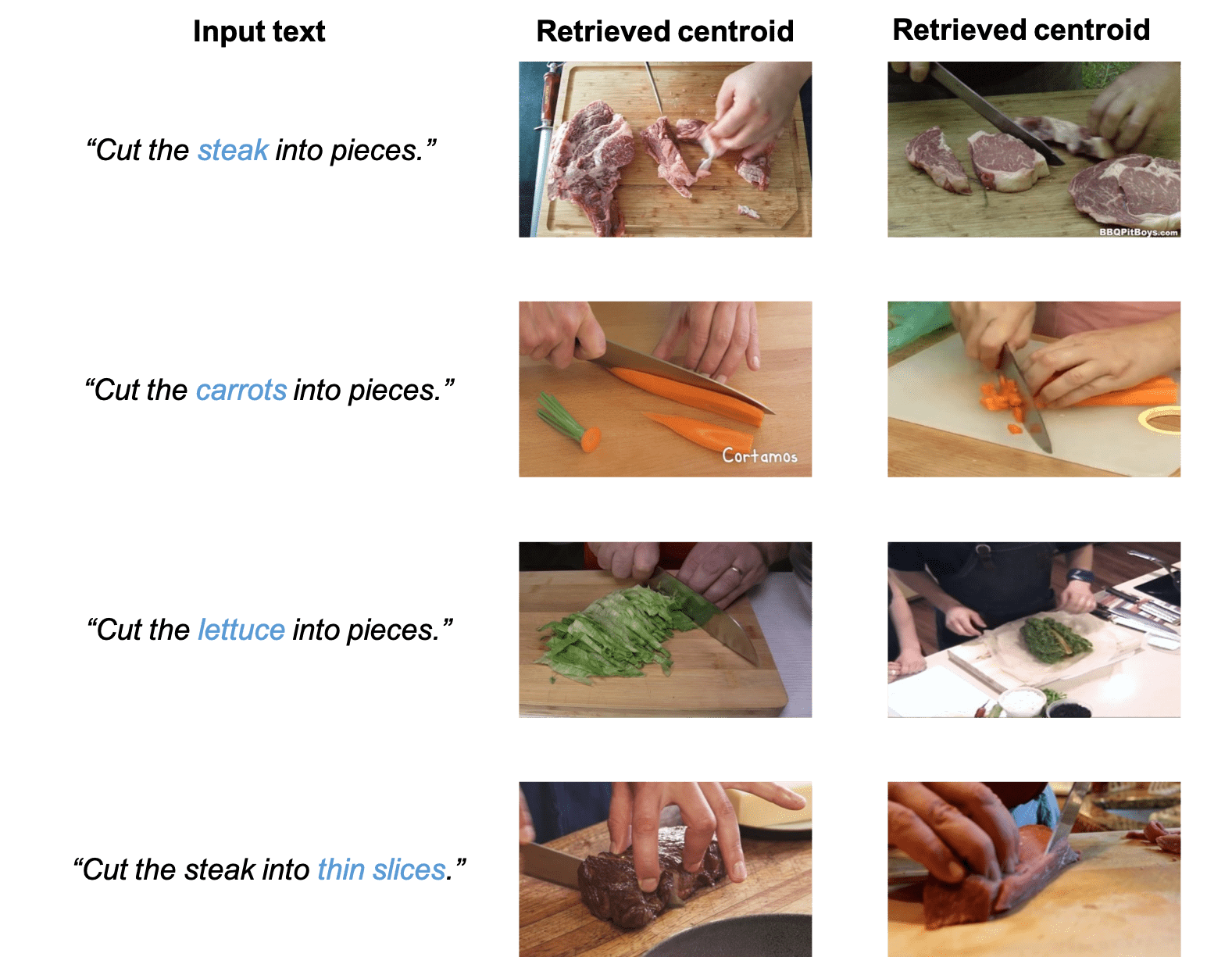}
    \caption{Visualizations for text to video prediction. In particular, we make small changes to the input text, and compare how the generated video tokens vary. We show top 2 retrieved video tokens for each text query.}
\end{figure*}

\end{document}